\def\year{2022}\relax
\documentclass[letterpaper]{article}
\usepackage{aaai22} % DO NOT CHANGE THIS
\usepackage{times} % DO NOT CHANGE THIS
\usepackage{helvet} % DO NOT CHANGE THIS
\usepackage{courier} % DO NOT CHANGE THIS
\usepackage[hyphens]{url} % DO NOT CHANGE THIS
\usepackage{graphicx} % DO NOT CHANGE THIS
\usepackage{xcolor}
\usepackage{graphicx} % DO NOT CHANGE THIS
\usepackage{natbib} % DO NOT CHANGE THIS
\usepackage{caption} % DO NOT CHANGE THIS
\DeclareCaptionStyle{ruled}%
{labelfont=normalfont,labelsep=colon,strut=off}
\usepackage{booktabs}
\usepackage{algorithm}
\usepackage{algorithmic}
\usepackage{newfloat}
\usepackage{listings}
\floatstyle{ruled}
\newfloat{listing}{tb}{lst}{}
\floatname{listing}{Listing}

\usepackage{multirow}

\graphicspath{ {./images/} }

\title{Mining Adverse Drug Reactions from \\ Unstructured Mediums at Scale}
\author{
    Hasham Ul Haq,
    Veysel Kocaman,
    David Talby
}
\affiliations{
    John Snow Labs Inc.\\
    16192 Coastal Highway\\
    Lewes, DE , USA 19958\\
    \{hasham, veysel, david\} @ johnsnowlabs.com\\
}

\begin{document}
\maketitle

\section{Abstract}
Adverse drug reactions / events (ADR/ADE) have a major impact on patient health and health care costs. Detecting ADR's as early as possible and sharing them with regulators, pharma companies, and healthcare providers can prevent morbidity and save many lives. While most ADR's are not reported via formal channels, they are often documented in a variety of unstructured conversations such as social media posts by patients, customer support call transcripts, or CRM notes of meetings between healthcare providers and pharma sales reps. In this paper, we propose a natural language processing (NLP) solution that detects ADR's in such unstructured free-text conversations, which improves on previous work in three ways. First, a new Named Entity Recognition (NER) model obtains new state-of-the-art accuracy for ADR and Drug entity extraction on the ADE, CADEC, and SMM4H benchmark datasets (\textbf{91.75\%}, \textbf{78.76\%}, and \textbf{83.41\%} F1 scores respectively). Second, two new Relation Extraction (RE) models are introduced - one based on BioBERT while the other utilizing crafted features over a Fully Connected Neural Network (FCNN) - are shown to perform on par with existing state-of-the-art models, and outperform them when trained with a supplementary clinician-annotated RE dataset. Third, a new text classification model, for deciding if a conversation includes an ADR, obtains new state-of-the-art accuracy on the CADEC dataset (\textbf{86.69\%} F1 score). The complete solution is implemented as a unified NLP pipeline in a production-grade library built on top of Apache Spark, making it natively scalable and able to process millions of batch or streaming records on commodity clusters.

\section{Introduction}
Adverse drug events are harmful side effects of drugs, comprising of allergic reactions, overdose response, and general unpleasant side effects. Approximately 2 million patients in the United States are affected each year by serious ADR's, resulting in roughly 100,000 fatalities \cite{leaman2010towards}, and making ADR's the fourth leading cause of death in the United States \cite{giacomini2007good}. Treatment related to ADR's has been estimated to cost \$136 billion each year in the United States alone \cite{van2006adverse}.

Finding all ADR's of a drug before it is marketed is not practical for several reasons. First, The number of human subjects going through clinical trials is often too small to detect rare ADR's. Second, many clinical trials are short-lasting while some ADR's take time to manifest. Third, some ADR's only show when a drug is taken together with other drugs, and not all drug-drug combinations can be tested during clinical trials. Fourth, drug repurposing or off-label usage can lead to unforeseen ADR's. As a result, detecting ADR's in drugs which are already being marketed is critical - a discipline known as postmarketing pharmacovigilance  \cite{mammi2013pharmacovigilance}.

Schemes which allow hospitals, clinicians, and patients to report ADR's have existed for many years, but only a fraction of events get reported through them. A meta-analysis of 37 studies from 12 countries found that the median rate of under-reporting was 94\% \cite{hazell2006under}. This led to work on mining ADR's from alternative sources, such as social media posts by patients or healthcare providers \cite{bollegala2018causality}. Outbreak of the COVID-19 pandemic has precipitated this trend of sharing such information \cite{cinelli2020covid}; The size, variety, and instantaneous nature of social media provides opportunities for real-time monitoring of ADRs \cite{sloane2015social}. Compared to traditional data source like research publications, this data is more challenging to process, as it is unstructured and contains noise in the form of jargon, abbreviations, misspellings, and complex sentence structures.

Recent advancements in Natural Language Processing (NLP) in the form of Transformers \cite{DBLP:journals/corr/VaswaniSPUJGKP17} based architectures like BERT \cite{DBLP:journals/corr/abs-1810-04805}, have significantly pushed the boundaries of NLP capabilities. There is an increasing trend of training large models on domain-specific data like BioBERT \cite{DBLP:journals/corr/abs-1901-08746}, and these methods have proven to achieve state-of-the-art (SOTA) results for document understanding and named entity recognition (NER). However, since these methods require significant computational resources during both training and inferring, it becomes impractical to apply them over large quantities of records in compute-restricted production environments.

Despite the growing interest and opportunities to process large quantities of data, models and software frameworks that can scale to leverage compute clusters are scarce. This restricts the ability to utilize available data from social media and other mediums - such as transcripts of customer service calls with patients, or CRM notes about sales and support discussions with clinicians - to their true potential. The availability of high volume, variety, and velocity of data presents the opportunity to develop NLP solutions that outperform the existing SOTA accuracy, while also being easily scalable and computationally efficient.

The purpose of this study is to illustrate how an end-to-end system, based on the Apache Spark ecosystem and comprising of novel NLP techniques, can be used to process large quantities of unstructured text to mine ADRs. This system has been implemented on a production-ready, widely deployed, and natively scalable library, thus capable of processing millions of records, in either batch or streaming modes. The unified NLP pipeline includes new models for the three required sub-tasks: classifying text to decide if it is an indication of an ADR, recognizing named entities for reactions and drugs, and linking adverse events with drugs. Following are the novel contributions of this paper: 

\begin{itemize}
 \item The first scalable end-to-end system for mining ADR's from unstructured text, including Document Classification, Named Entity Recognition, and Relation Extraction Models within a unified NLP pipeline.
 \item New NER model for extracting reactions and drugs, whose accuracy outperforms previous SOTA models on public datasets for this task.
 \item New Relation Extraction models for linking reactions and drugs, which outperform previous SOTA models when trained with additional data that was annotated by clinicians as part of this effort.
 \item New text classification model for deciding if a piece of text reports an ADR, whose accuracy outperforms previous SOTA models.
 \item Studying the utility of using non-contextual lightweight embeddings \cite{DBLP:journals/corr/abs-1301-3781} like GloVe \cite{pennington-etal-2014-glove} instead of memory-intensive contextual embeddings like BioBERT for these tasks, by comparing training times and accuracy improvements.
 \item Detailed analysis of all the solution components and datasets, explaining how its modular structure can be customized to different data sources and runtimes.
\end{itemize}

\section{Related Work}

The extraction of ADRs from unstructured text has received growing attention in the past few years due to wide-spread adoption of Electronic Medical Records (EMR), and ever-increasing number of users on social media who share their experiences. Existing work comprises of significant contributions in both, novelty in information extraction methodologies, as well as availability of relevant pre-annotated datasets containing annotations for a variety of subtasks.

The problem of ADR extraction gained visibility with the introduction of challenges like Social Media Mining for Healthcare (SMM4H) \cite{ws-2019-social} and National Clinical NLP Challenges (n2c2) \cite{henry20202018}, which provide pre-annotated datasets for researchers to compete on. Other significant contributions for data collection include \cite{GURULINGAPPA2012885} which used the Pubmed corpus to develop the ADE corpus benchmark dataset, covering Classification, NER, and RE annotations for extracting and relating ADRs and drugs respectively. Another work \cite{article} produced an NER dataset (CADEC) by collecting and annotating reviews and comments from forums.

Identification of text containing ADRs is formulated as a text classification problem for which different techniques have been applied. \cite{huynh-etal-2016-adverse} used different variations of Convolutional Neural Network (CNN) (e.g., CNN, CRNN, CNNA) to identify tweets containing ADRs on the twitter dataset. More elaborate techniques like fine-tuning of BERT models have been applied for text classification as well \cite{kayastha-etal-2021-bert}.

A standard method of formulating the extraction of drugs and ADR mentions is NER, for which, a number of architectures have been proposed. One of the classical approach is to use a BiLSTM \cite{GRAVES2005602} architecture with Conditional Random Fields (CRF) as used by \cite{stanovsky-etal-2017-recognizing}. This method is a shallow network that relies on word embeddings and part of speech tags to classify each token to extract ADR mentions. \cite{DBLP:journals/corr/abs-2003-09288} also added character level embeddings to the same architecture to incorporate spelling features, and enriched the training dataset by annotating additional data from DBpedia to achieve SOTA results on the CADEC dataset, demonstrating the benefits of using additional data. Similar to our approach, they also built an extensive training framework over multiple nodes.

Relating ADR mentions with the drugs is formulated as a relation extraction (RE) task, which comprises of creation and classification of relations between entities \cite{haq2021deeper}. Classical RE methods like \cite{10.1093/bioinformatics/btl616} use lexical rules based on dependency parsing tree of the document. The introduction of transformer allowed for more context-aware solutions like feeding entity spans and document to transformers to predict relations \cite{DBLP:journals/corr/abs-1906-03158}. Recently, more elaborate approaches like joint learning of both NER and RE have proved to be more beneficial. For example, \cite{DBLP:journals/corr/abs-2002-06424} used a single base network to generate joint features, while using separate BiRNN layers for both NER and RE, and creating skip connections between the NER and RE BiRNN layers to achieve SOTA performance on RE.

While existing work has focused on pushing the boundaries for accuracy, little work is done to build a framework that can process large quantities of data from social media with accuracy. To achieve this, we develop separate architectures for all three tasks, and place them in a single pipeline, allowing us to maintain a modular structure to develop and test each component separately, while sharing common components (e.g., tokenization and embedding generation) for scalability. 

\section{Approach}
\begin{figure}[h!]
  \includegraphics[width=0.5\textwidth]{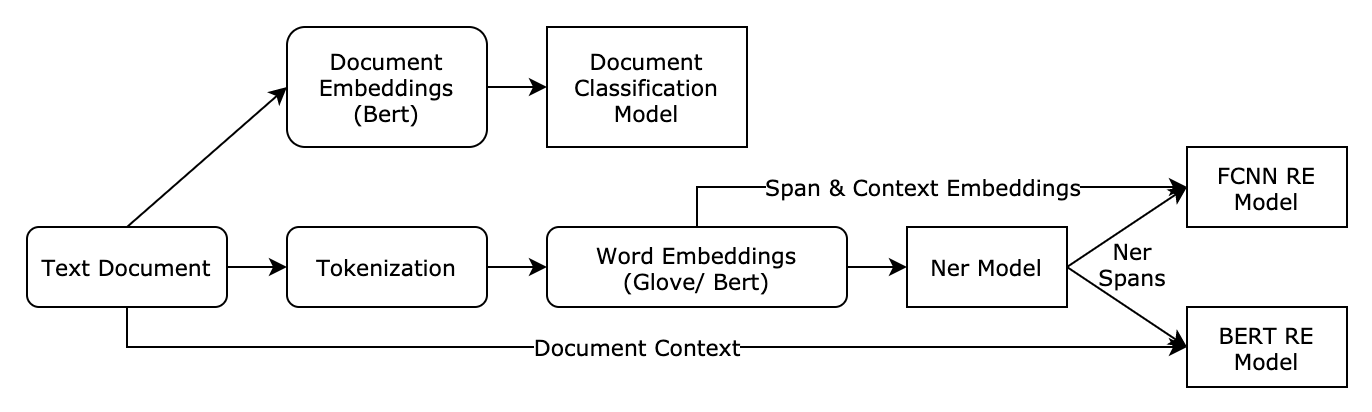}
  \caption{Overview of the complete architecture. All the components are sequentially placed in a single pipeline. Arrows represent output of one stage as input to the next stage.}
  \label{fig:basic_pp}
\end{figure}

We divide the problem into three main tasks; Document Classification, Named Entity Recognition and Relation Extraction, and draw distinct solutions for each one of them for scalability. Since NER plays the most important role of identifying entity spans, we place all components in a single pipeline for an end-to-end solution. Figure \ref{fig:basic_pp} explains the complete pipeline using Apache Spark framework.

As illustrated in the system diagram in Figure \ref{fig:basic_pp}, Relation Extraction is heavily dependent on the NER model, as the latter provides relevant entity chunks which form basic inputs of the RE model. Since NER requires token level embeddings, we test with different types of embeddings; namely GLoVe \cite{pennington-etal-2014-glove} and BERT \cite{DBLP:journals/corr/abs-1810-04805} based embeddings. This modular approach helps us keep the NER and RE architecture static while experimenting with different embedding types to analyse accuracy and performance differences. Given the nature of the data, we trained 200-dimension GLoVe embeddings on Pubmed and MIMIC datasets. For BERT embeddings we utilize the work by \cite{DBLP:journals/corr/abs-1901-08746}, namely BioBERT. In general, BERT  embeddings provide more useful information due to being context-aware and better handling of out of vocabulary (OOV) tokens.

\subsection{Classification}
%\vk{TO-DO section - will add more stuff during the day}
To be able to process large volume of data, the text classification model needs to be scalable, and accurate, as it is used to filter out documents, reviews, and tweets that do not contain any indication of adverse event. To achieve this, we use a FCNN model that does not require  hand-crafted features, and relies on a single embedding vector for classification. Given the conversational nature of social media text, we can utilise the entire document to get efficient embeddings (with little text clipping in case of BioBERT embeddings) that we directly feed to the classifier model. Since there is only a single feature vector as input to the model, we test multiple embedding techniques to analyse performance.

\subsection{Named Entity Recognition}
\begin{figure}[h!]
  \includegraphics[width=0.5\textwidth]{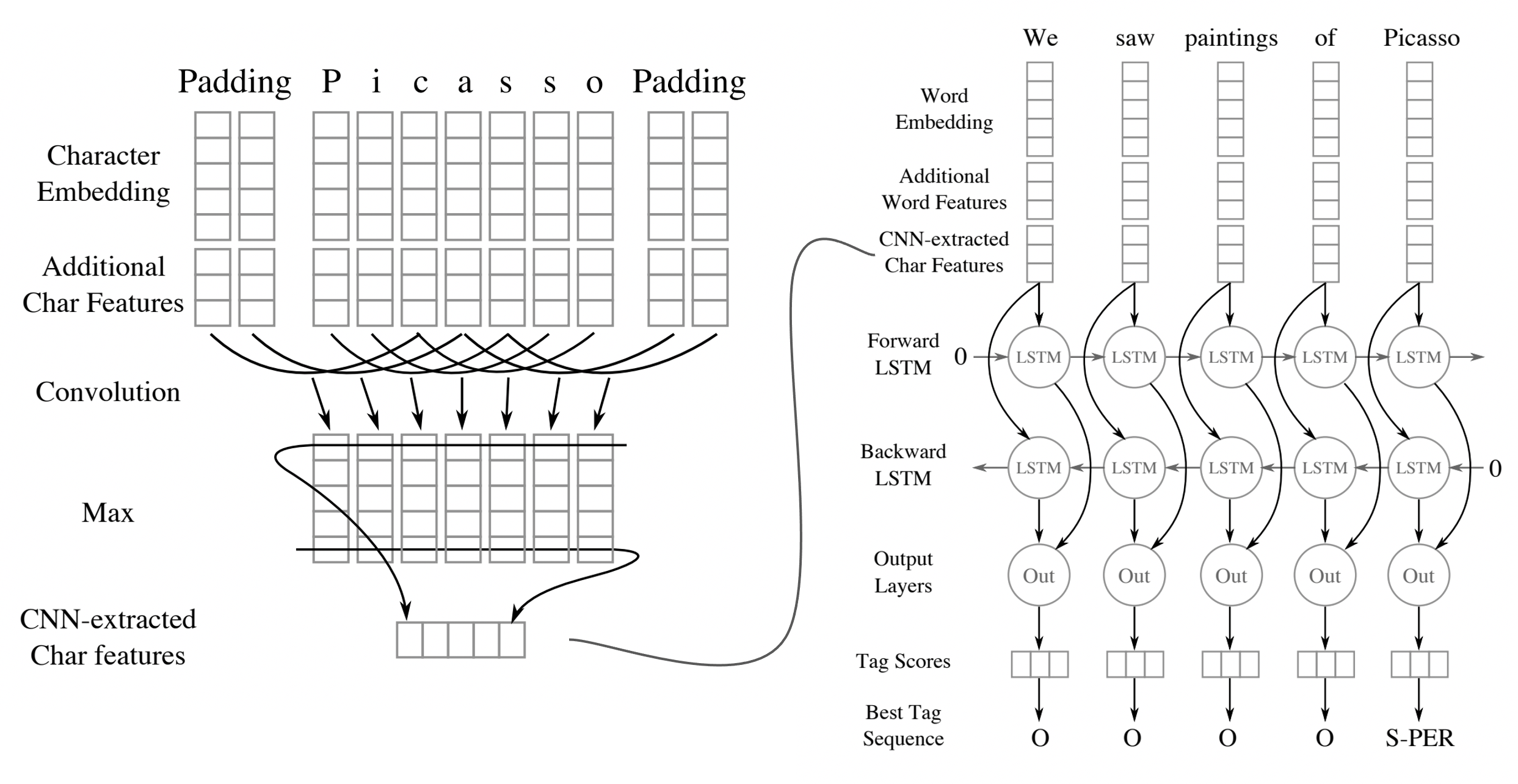}
  \caption{Proposed NER architecture, as explained in  \cite{DBLP:journals/corr/abs-2011-06315}.}
  \label{fig:ner_model}
\end{figure}

To extract ADR and other entities from text, we use our class-leading NER architecture, called BiLSTM-CNN-Char. We build our NER model by taking the work of \cite{DBLP:journals/corr/ChiuN15} as the base model, and made a few changes in the architecture according to our testing; removing lexical features like POS tags, and introducing new character level features. We used 1D convolution layer comprising of 25 filters having kernel size 3 to generate token feature maps that encapsulate information like spelling and casing. These additional features proved highly useful while dealing with spelling mistakes, as well as out-of-vocabulary tokens. We also updated the architecture by using BlockFused-LSTM cells in our implementation for increased speed. Figure \ref{fig:ner_model} explains the architecture of our NER model.

\subsection{Relation Extraction}

We treat Relation Extraction (RE) as a binary classification problem where each example is a pair of drug and ADR mentions in a given context, and develop two novel solutions; the first one comprising of a simpler FCNN architecture for speed, and the second one based on the BioBERT architecture for accuracy. We experiment both approaches and compare their results. 

\begin{figure}[h!]
  \includegraphics[width=0.45\textwidth]{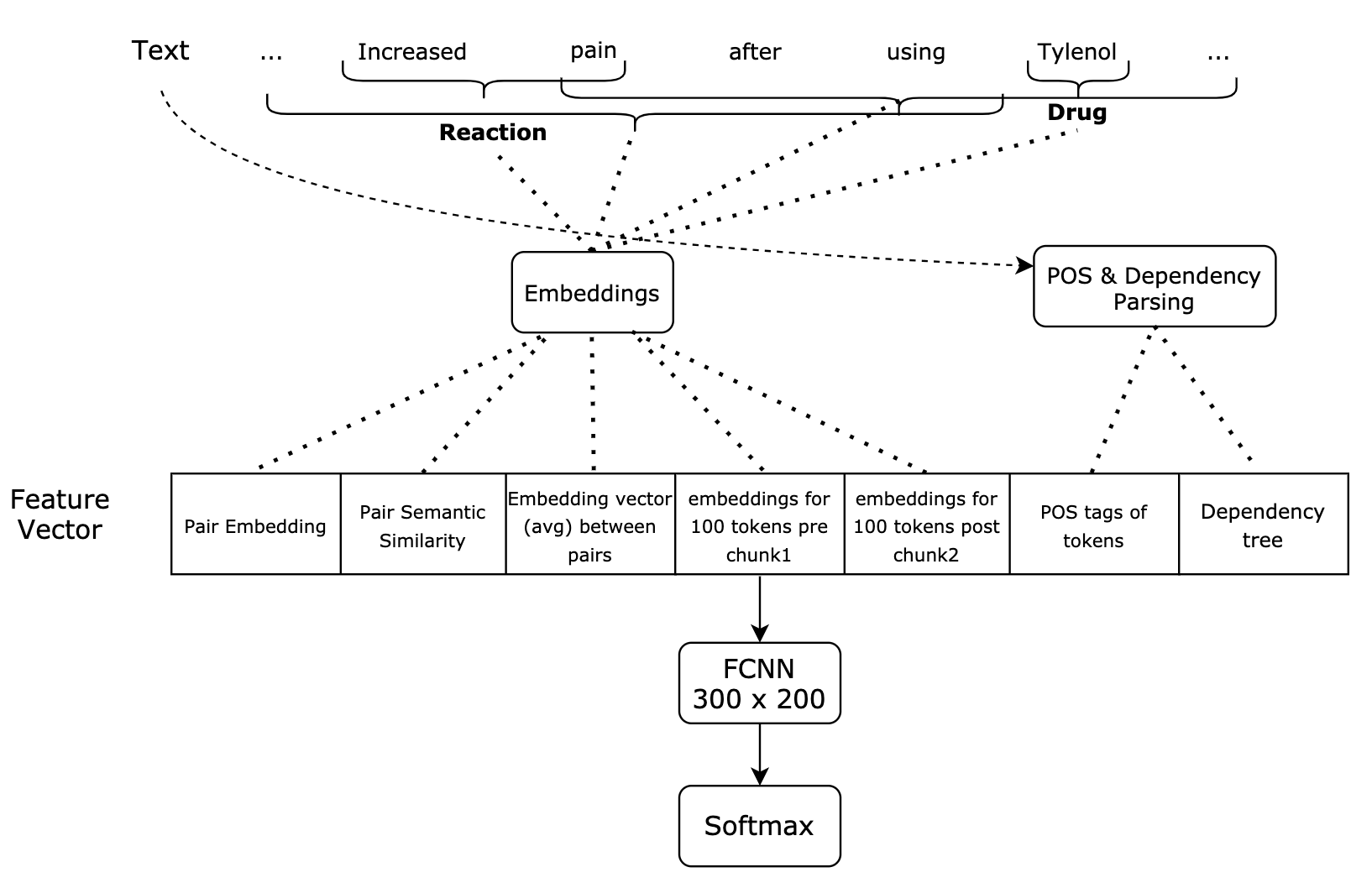}
  \caption{Overview of the first Relation Extraction model. All the features are vertically stacked in a single feature vector. The feature vector is kept dynamic with additional padding for compatibility across different embedding sizes, and complex dependency structures.}
  \label{fig:re_model}
\end{figure}

For our first RE solution we rely on entity spans and types identified by the NER model to develop distinct features to feed to an FCNN for classification. At first we generate pairs of adverse event and drug entities, and then generate custom features for each pair. These features include semantic similarity of the entities, syntactic distance of the two entities, dependency structure of the entire document, embedding vectors of the entity spans, as well as embedding vectors for 100 tokens within the vicinity of each entity. Figure \ref{fig:re_model} explains our model architecture in detail. We then concatenate these features and feed them to fully connected layers with leaky relu activation. We also use batch normalisation after each affine transformation before feeding to the final softmax layer with cross-entropy loss function. We use softmax cross-entropy instead of binary cross-entropy loss to keep the architecture flexible for scaling on datasets having multiple relation types.

Our second solution focuses on a higher accuracy, as well as exploration of relations across long documents, and is based on \cite{DBLP:journals/corr/abs-1906-03158}. In our experiment we take checkpoints from the BioBERT model and train an end-to-end model for relation extraction. Similar to our first solution, we rely on entity spans and use the entire document as context string while training the model. The original paper used sequence length of 128 tokens for the context string, which we keep constant, and instead experiment with the context string, additional data, and fine-tuning techniques.

\section{Experimental Setup}

\subsection{Datasets}

We test our models on three benchmark datasets; SMM4H NER challenge \cite{ws-2019-social}, ADE Corpus \cite{GURULINGAPPA2012885} and CADEC \cite{article}. The SMM4H NER challenge is a yearly challenge based on annotated twitter data. As this dataset is entirely based on tweets, it forms an ideal testing bed to test our model's performance on real world data. The ADE Corpus dataset is a benchmark dataset for classification, NER and RE tasks, while the CADEC dataset is primarily used for classification and NER benchmarks only. Keeping consistency with existing work, as well as aligning with our primary goal of extracting ADRs and related drugs, we keep two entities in all datasets; ADE and Drug. Details of the NER datasets can be found in Table \ref{tab:ner_data_stats}

\begin{table}[ht]
\centering
\begin{tabular}{lccc}
\toprule
 \textbf{Dataset} & \textbf{\# Sentence} & \textbf{\# Token} & \textbf{\# Entity tags}\\
\midrule
 \multirow{2}{*}{ADE Corpus} & \multirow{2}{*}{4272} & \multirow{2}{*}{86865} & ADE: 12264\\
 &  &  & Drug: 5544\\ 
 \multirow{2}{*}{CADEC} & \multirow{2}{*}{7597} & \multirow{2}{*}{121656} & ADE: 15903\\
  &  &  & Drug: 2032\\
 \multirow{2}{*}{SMM4H} & \multirow{2}{*}{2253} & \multirow{2}{*}{42175} & ADE: 3575\\
  &  &  & Drug: 1586\\
\bottomrule
\end{tabular}
\caption{Statistics of the benchmark NER datasets.}
\label{tab:ner_data_stats}
\end{table}

Since we treat the RE problem as binary classification, we need positive relations as well as negative relations to train the model. Positive relations are defined if the drug and reaction entities are related in the context, while negative relations comprise of drugs that are not responsible for a particular reaction. This relation can be formulated as below: \[P(Drug|ADE)\] From the ADE dataset, we can sample negative relations by subtracting annotated drug-reaction pairs from all drug-reaction pairs in the same document. Table \ref{tab:re_data_stats} shows data distribution of the standard and enriched RE datasets. 

\begin{table}[ht]
\centering
\begin{tabular}{lcc}
\toprule
 \textbf{Dataset} & \textbf{\# Positive Rel.} & \textbf{\# Negative Rel.}\\
\midrule
 ADE Corpus & 6821 & 183 \\ 
 ADE with n2c2 & 12929 & 8935\\
\bottomrule
\end{tabular}
\caption{Statistics of the RE datasets used to train and validate the models. ADE Corpus is the standard dataset, which is then enriched with n2c2 data for more robust performance.}
\label{tab:re_data_stats}
\end{table}

The standard ADE Corpus does not have sufficient negative relations, raising the issue of class imbalance. To address this, we sampled and annotated ~2000 notes from 2018 n2c2 shared task on ADE and medication extraction in EHRs dataset \cite{henry20202018}, to create a supplementary dataset for relations. We keep the same entities (i.e., Drug and ADE) while annotating to align with our benchmark datasets. Also, to keep human bias at a minimum, we don't annotate entity spans; rather we use existing NER annotations to generate a dataset comprising of Drug and ADE pairs, and only classify each relation based on their context.

Following previous work, we evaluate the models using 10-fold cross validation, and report macro and micro averaged precision, recall, and F1 scores. Exact experimental and evaluation settings for each stage are described below.

\subsection{Experiments}

\begin{itemize}
    \item Keeping the Classification model architecture constant, we test two methods of embedding generation; For the first experiment we generated token level GLoVe and BioBERT embeddings for each token and averaged them to generate document embedding. This method did not produce accurate embeddings, resulting in lower F1 scores. Our second experiment utilised Sentence Bert Embeddings \cite{DBLP:journals/corr/abs-1908-10084} trained on the Pubmed dataset, namely Sentence BioBERT, which we further pretrained on the MedNLI dataset \cite{shivade2019mednli} for better performance. We thoroughly test the performance of each embedding generation approach, and report the findings in Table \ref{tab:cls_metrics}.
    
%%%
%%% cls table
%%%
\begin{table*}[ht]
\centering
\begin{tabular}{lcccccccccc}
\toprule
\multirow{2}{*}{\textbf{Dataset}} & \multicolumn{3}{c}{\textbf{GLoVe (Avg.) Embeddings}} & \multicolumn{3}{c}{\textbf{BERT (Avg.) Embeddings}} & \multicolumn{3}{c}{\textbf{BERT Sentence Embeddings}} & \textbf{SOTA}\\
%\cline{2-11}
\multirow{2}{*}{} & Precision & Recall & F1 & Precision & Recall & F1 & Precision & Recall & F1 & F1 \\
\midrule
% & & & & & & & & & \\
\multirow{2}{*}{ADE} & 75.96 & 79.53 & 76.86 & 76.91 & 84.96 & 79.37 & 87.41 & 84.72 & \textbf{85.96} & \textbf{87.0} \\
 & 86.84 & 81.22 & 83.43 & 88.13 & 84.38 & 85.38 & 90.97 & 91.20 & 91.03 & \\
 & & & & & & & & & \\
\multirow{2}{*}{CADEC} & 85.29 & 84.24 & 84.71 & 86.50 & 86.11 & 86.30 & 87.13 & 86.32 & \textbf{86.69} & \textbf{81.5}\\
 & 85.99 & 86.10 & 86.0 & 87.38 & 87.43  & 87.40 & 87.78 & 87.86 & 87.79 &  \\
\bottomrule
\end{tabular}
\caption{Classification Metrics on benchmark datasets. For each dataset, Macro and Micro averaged scores are displayed on first and second row respectively. SOTA metrics for ADE and CADEC datasets are obtained from \cite{huynh-etal-2016-adverse} and \cite{sota_cadec_cls} respectively.}
\label{tab:cls_metrics}
\end{table*}

\begin{table*}
\centering
\scalebox{0.95}{
\begin{tabular}{ lccccccccccccc}
\toprule
\multirow{3}{*}{Dataset} &\multicolumn{6}{c}{\textbf{GLoVe Embeddings}} &\multicolumn{6}{c}{\textbf{BERT Embeddings}} & \textbf{SOTA}\\
%\cline{2-14} 

\multirow{2}{*}{} &\multicolumn{2}{c}{Precision} &\multicolumn{2}{c}{Recall} &\multicolumn{2}{c}{F1} &\multicolumn{2}{c}{Precision} &\multicolumn{2}{c}{Recall} &\multicolumn{2}{c}{F1} & F1 \\

\multicolumn{1}{c}{} & \multicolumn{1}{c}{strict} & \multicolumn{1}{c}{relax} & \multicolumn{1}{c}{strict} & \multicolumn{1}{c}{relax}& \multicolumn{1}{c}{strict} & \multicolumn{1}{c}{relax} & \multicolumn{1}{c}{strict} & \multicolumn{1}{c}{relax} & \multicolumn{1}{c}{strict} & \multicolumn{1}{c}{relax}& \multicolumn{1}{c}{strict} & \multicolumn{1}{c}{relax} \\
\midrule
\multirow{2}{*}{ADE} & 88.32 & 93.77 & 89.26 & 94.80 & 88.78 & 94.27 & 90.0 & 94.47 & 93.56 & 98.22 & \textbf{91.75} & 96.31 & \textbf{91.3}\\
 & 87.81 & 93.59 & 88.81 & 94.66 & 88.30 & 94.12 & 89.6 & 94.37 & 93.18 & 98.13 & 91.36 & 96.21 \\
 & & & & & & & & & & & & \\
\multirow{2}{*}{CADEC} & 78.14 & 89.04 & 77.14 & 88.01 & 77.62 & 88.50 & 78.53 & 88.63 & 79.03 & 89.32 & \textbf{78.76} & 88.95 & \textbf{71.9}\\
 & 71.87 & 86.36 & 71.67 & 86.13 & 71.75 & 86.23 & 72.38 & 86.14 & 73.64 & 87.66 & 72.99 & 86.88  \\
 & & & & & & & & & & & & \\
 \multirow{2}{*}{SMM4H} & 81.43 & 90.33 & 72.17 & 78.51 & 76.01 & 83.41 & 78.5 & 86.76 & 75.23 & 82.42 & \textbf{76.73} & 84.41 & \textbf{67.81}\\
 & 83.66 & 91.34 & 71.31 & 77.86 & 76.99 & 84.06 & 79.13 & 87.09 & 74.33 & 81.81 & 76.65 & 84.36\\
\bottomrule
\end{tabular}}
\caption{NER metrics on benchmark datasets. For each dataset, macro and micro averaged scores are displayed on first and second row respectively. SOTA metrics for ADE, CADEC, and SMM4H are obtained from \cite{yan2021partition}, \cite{stanovsky-etal-2017-recognizing}, and \cite{DBLP:journals/corr/abs-2003-09288} respectively, and are macro-averaged.}
\label{tab:ner_metrics}
\end{table*}

\begin{table}[ht]
\centering
\scalebox{0.93}{
\begin{tabular}{lcccccc}
\toprule
 \multirow{2}{*}{\textbf{Dataset}} & \multicolumn{3}{c}{\textbf{BERT}} & \multicolumn{3}{c}{\textbf{GLoVe}}\\
 & Train & Infer & F1 & Train & Infer & F1\\
\midrule
 ADE & 1980s & 329s & 91.75 & 1417s & 22s & 88.78\\ 
 CADEC & 2475s & 351s & 78.76 & 1929s & 26s & 77.62\\
 SMM4H & 860s & 136s & 76.73 & 635s & 11s & 76.01\\
\bottomrule
\end{tabular}}
\caption{Training and inference time (in seconds) taken by the NER model on each dataset using different token embeddings with respect to overal performance on test set. Epoch count was kept constant for all datasets while training. The experiment was performed on an 8-core machine having 64gb memory.}
\label{tab:time_stats}
\end{table}
    
    \item For the NER model we align all the datasets in the standard CoNLL format and use IOB (Inside, Outside, Beginning) tagging scheme. We tested other tagging schemes as well, like BIOES, and found IOB to be the simplest and best performing scheme. Since our NER architecture requires word embeddings, we experiment with two types of embeddings. For the GLoVe embeddings, we use our 200-dimension embeddings, and leverage BioBERT embeddings for contextual embeddings.
    
    For thorough analysis of the NER model, we evaluate the results using both strict and relax approach. Under strict evaluation a label is considered as correct if the starting and ending tags exactly match with the gold labels, while under relax evaluation only an overlap between annotations is considered. Consequently, the ‘O’ tag is not included in the calculation. Hyperparameter values, and training code is explained in Appendix A \& B.
    
    \item For training RE models, we use standard NER spans and binary labels. For our base RE model we use 200-dimensional token-level GLoVe embeddings - the same embeddings we use for our base NER model.
    
    For our BERT based RE model, we don't use any explicit embeddings as the BERT model is trained in an end-to-end fashion. We do specify details of entity spans like starting, ending indices, entity types, and the context in between. The context is generally the entire document, but since the model architecture has a 128 token limit, we create context text by taking text in between the entities, and found this method to be more accurate.
    
    We also test a hypothesis that fine-tuning BioBERT model on similar Relation Extraction tasks would increase the overall performance on the benchmark datasets. To test this hypothesis, we train an end-to-end RE model on Disease and Drug datasets like the 2010 i2b2 challenge \cite{uzuner20112010} and saved it. We then use the same weights while discarding the final layers, and retrain the model on the benchmark dataset. Since the base model is trained on a similar taxonomy, the convergence was much faster, while being less prone to over-fitting.
    
     For Hyperparameter tuning we utilize the development set and use random search. Exact hyperparameter values, and the search space for all the models can be found in Appendix A.
    
\end{itemize}

\subsection{Results}

\begin{table*}
\centering
\begin{tabular}{lccccccc}
\toprule
\multirow{2}{*}{\textbf{Dataset}} &\multicolumn{3}{c}{\textbf{Base (FCNN) RE}} &\multicolumn{3}{c}{\textbf{BERT RE}} & \textbf{SOTA}\\
 & Precision & Recall & F1 & Precision & Recall & F1 & F1\\
\midrule
ADE Corpus & 69.11 & 86.22 & \textbf{74.70} & 81.31 & 79.03 & \textbf{80.10} & \textbf{83.74} \\
ADE Enriched with n2c2 & 89.01 & 89.44 & \textbf{89.22} & 89.19 & 90.93 & \textbf{90.02} & \\
\bottomrule
\end{tabular}
\caption{Relation Extraction performance on the ADE benchmark dataset. The test set was kept standard for a fair comparison, and all scores are macro-averaged due to high class imbalance. SOTA metrics for RE on ADE corpus as reported by \cite{DBLP:journals/corr/abs-2002-06424}}
\label{tab:re_metrics}
\end{table*}

\begin{table*}
\centering
\begin{tabular}{| c | c | c | c | c |}
\hline
\textbf{Document} & \textbf{Class} & \textbf{ADE Entity} & \textbf{Drug Entity} & \textbf{relation}\\
\hline
\multirow{2}{*}{I feel a bit drowsy \& have a little blurred vision after taking insulin.} & \multirow{2}{*}{ADE} & drowsy & insulin & Positive\\
&  & blurred vision & insulin & Positive\\
\hline
\multirow{2}{*}{@yho fluvastatin gave me cramps, but lipitor suits me!} & \multirow{2}{*}{ADE} & cramps & fluvastatin & Positive\\
&  & cramps & lipitor & Negative\\
\hline
I just took advil and haven't had any gastric problems so far. & NEG & -  & -  & - \\
\hline
\end{tabular}
\caption{Full pipeline results on sample texts. Documents having indication for ADR are classified as ADE, while positive relations represent causality between two entities (Drug and ADE). The last example is classified as negative - meaning it does not contain any ADE indication, so we don't process it further.}
\label{tab:examples}
\end{table*}

Despite using a shallow architecture for classification, we achieved metrics that are on-par with SOTA metrics by using more accurate Sentence Bert Embeddings, as shown in Table \ref{tab:cls_metrics}. While the performance difference between BioBERT and GLoVe embeddings is minor on the CADEC dataset, the difference is more prominent on the ADE dataset. This is primarily because of the complex intrinsic nature of biomedical text, where averaging token (GLoVe) embeddings does not efficiently capture the context of complex sentence structures.

Our NER architecture acheives new SOTA metrics on SMM4H, ADE, and CADEC NER datasets using contextual BioBERT embeddings as shown in Table \ref{tab:ner_metrics}. Since the NER model was kept constant during the experiment, and we tuned the hyper parameters for each experiment, the performance difference between embedding types can be attributed to the word embeddings alone. Being able to incorporate contextual information with attention mechanism, BioBERT embeddings peformed better than non-contextual GLoVe embeddings. However, it is worth noticing that the performance difference between the two is in a margin of ~1-2\%, proving that domain specific GLoVe embeddings can provide comparable performance while requiring significantly less memory and computational resources. Table \ref{tab:time_stats} provides side-by-side comparison of time and accuracy differences while using different embedding types. On average, the GLoVe embeddings are ~30\% faster compared to BioBERT embeddings during training, and more than 5x faster during inference, while being on-par in terms of f1 score. 

Our RE solutions perform on-par with existing SOTA systems, while being scalable and requiring less memory to train and test. The introduction of the extra data greatly improved results, enabling us to achieve SOTA on benchmark datasets as shown in Table \ref{tab:re_metrics}. While the more heavy BioBERT model outperformed our proposed RE model on the limited and imbalanced ADE dataset, the performance difference becomes diminutive when more data is added to the training data.

\begin{figure*}[h!]
  \includegraphics[width=\textwidth]{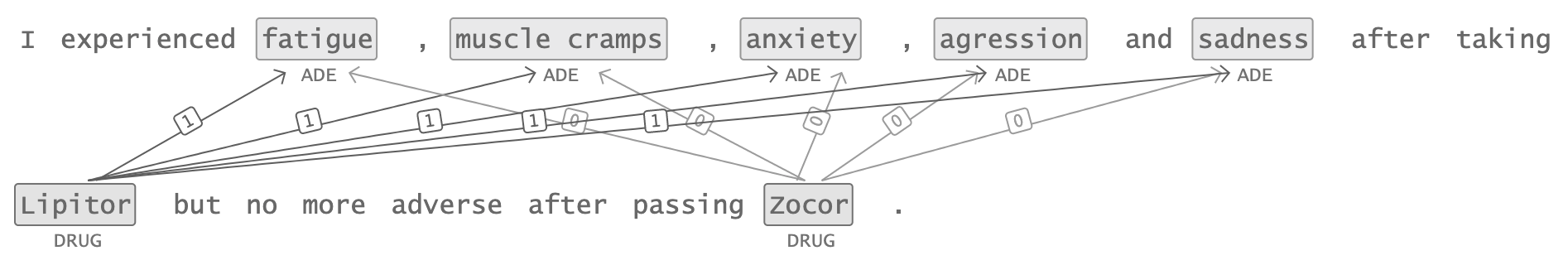}
  \caption{Visualization of Entity Recognition and Relation Extraction results on a sample text. 1 denotes positive relation and 0 denotes negative relation (not related).}
  \label{fig:example_1}
\end{figure*}

Sample output and visualization of NER and RE results can be seen in Table \ref{tab:examples} and Figure \ref{fig:example_1}.

\section{Conclusion}
Despite the growing need and explosion of useful data for pharmacovigilance, there is a severe deficiency of production-ready NLP systems that can process millions of records while being accurate and versatile.

In this study we address the problem by introducing novel solutions for Classification, NER, and RE while leveraging the Spark ecosystem and contemplating on accuracy, scalability, and versatility. For which we explain how we build a modular structure comprising of different embedding types, a classification and NER model, and two approaches for RE. We trained custom GLoVe embeddings model on domain-specific dataset, and compare its performance to SOTA BioBERT embeddings. We show through extensive testing that our text classification model, for deciding if a conversation includes an ADR, obtains new state-of-the-art accuracy on the CADEC dataset (86.69\% F1 score). Our proposed NER architecture achieves SOTA results on multiple benchmark datasets. Namely, our proposed NER models obtain new state-of-the-art accuracy for ADR and Drug entity extraction on the ADE, CADEC, and SMM4H benchmark datasets (91.75\%, 78.76\%, and 83.41\% F1 scores respectively). Then we explain two different architectures for RE, one based on BioBERT while the other utilizing crafted features over a FCNN, test them individually, and show that a simpler RE architecture with bespoke features performed on-par with more sophisticated BERT solution. To improve our RE model, we built a new dataset by manual annotations, and achieved higher metrics on the RE test datasets.

Furthermore, we performed speed benchmarks to compare efficiency of two distinct embedding generation models to determine the ideal choice for deploying such solutions to process large quantities of data. In general, most pharmaceutical companies run on-premise servers which are geared towards general computation and do not utilise hardware acceleration like GPUs for running heavy models; In such cases where infrastructure is not mature enough to handle heavy models, lightweight glove-based models are a compelling alternative to BERT-based models, as they offer comparable performance while being memory and CPU efficient.

Finally, we implement all these algorithms in Apache Spark ecosystem for scalability, and shipped in a production grade NLP library: Spark NLP.

\appendix
\section{A. Hyperparameter Settings}

The following parameters provided best results on the \textbf{classification} development set (values within the parenthesis represent the
parameter ranges tested):
\begin{itemize}
    \item Dropout rate: 0.2 (0.2, 0.7)
    \item Batch size: 8 (4, 256)
    \item Learning rate: 0.0003 (0.01, 0.0003)
    \item Epoch: 25-30 (10, 100)
    \item Optimizer: Adam
    \item Learning rate decay coefficient (po) (real learning rate =
lr / (1 + po * epoch) : 0.005 (0.001, 0.01))
\end{itemize}

The following parameters provided best results on the \textbf{NER} development set (Values within the parenthesis represent the parameter ranges tested):
\begin{itemize}
        \item LSTM state size: 200 (200, 250)
        \item Dropout rate: 0.5 (0.3, 0.7)
        \item Batch size: 8 (4, 256)
        \item Learning rate: 0.001 (0.01, 0.0003)
        \item Epoch: 25-35 (10, 100)
        \item Optimizer: Adam
        \item Learning rate decay coefficient (po) (real learning rate =
    lr / (1 + po * epoch) Smith [2018] : 0.005 (0.001, 0.01))
\end{itemize}

The following parameters provided best results on the \textbf{RE} development set (Values within the parenthesis represent the parameter ranges tested):
\begin{itemize}
        \item Dropout rate: 0.5 (0.3, 0.7)
        \item Batch size: 8 (4, 256)
        \item Learning rate: 0.0001 (0.01, 0.0003)
        \item Epoch: 4-BERT (1-10), 50-FCNN (10-100) 
\end{itemize}

\section{B. Training Code}
Code for training an RE model is provided as a google colab notebook \cite{re_training_code}. 

\newpage
\bibliography{references}

% \begin{quote}
% \begin{small}
% \bibliography{aaai22}
% \end{small}
% \end{quote}
 
\end{document}